\documentclass{article}
\pdfoutput=1

\usepackage{arxiv}

\usepackage[utf8]{inputenc} 
\usepackage[T1]{fontenc}    
\usepackage{hyperref}       
\usepackage{url}            
\usepackage{booktabs}       
\usepackage{nicefrac}       
\usepackage{microtype}      
\usepackage{graphicx}

\usepackage{doi}

\usepackage{amsmath,amsfonts}
\usepackage{algorithmic}
\usepackage{algorithm}
\usepackage{array}
\usepackage[caption=false,font=normalsize,labelfont=sf,textfont=sf]{subfig}
\usepackage{textcomp}
\usepackage{stfloats}
\usepackage{url}
\usepackage{verbatim}
\usepackage{graphicx}
\usepackage{cite}

\usepackage{comment}
\usepackage{color}
\usepackage{multirow}
\usepackage{tabularx}
\usepackage{booktabs}

\title{Motion-Based Sign Language Video Summarization using Curvature and Torsion}

\author{ \href{https://orcid.org/0009-0000-7807-4219}{\includegraphics[scale=0.06]{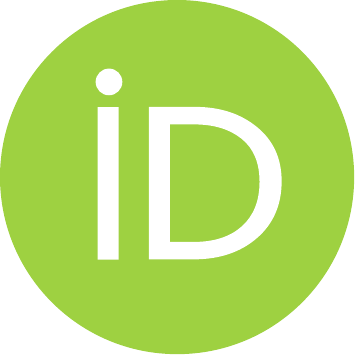}\hspace{1mm}Evangelos G. Sartinas}
\\
	\And
	\href{https://orcid.org/0000-0002-9627-0640}{\includegraphics[scale=0.06]{orcid.pdf}\hspace{1mm}Emmanouil Z. Psarakis} \\
    \And
	\href{https://orcid.org/0000-0003-3325-1247}{\includegraphics[scale=0.06]{orcid.pdf}\hspace{1mm}Dimitrios I. Kosmopoulos} \\
    \And
    \\
	Department of Computer Engineering and Informatics\\
	University of Patras, Patras, Greece 26504\\
	\texttt{\{sartinas, psarakis, dkosmo\}@ceid.upatras.gr} \\
}

\date{}

\renewcommand{\shorttitle}{Motion-Based Sign Language Video Summarization using Curvature and Torsion}

\hypersetup{
pdftitle={Motion-Based Sign Language Video Summarization using Curvature and Torsion},
pdfsubject={SL Video Summarization},
pdfauthor={Evangelos G.~Sartinas, Emmanouil Z.~Psarakis, Dimitrios I.~Kosmopoulos},
pdfkeywords={video summarization, curvature, sign language, Frenet-Serret frame, torsion},
}

\begin{document}
\maketitle

\begin{abstract}
An interesting problem in many video-based applications is the generation of short synopses by selecting the most informative frames, a procedure which is known as video summarization. For sign language videos the benefits of using the $t$-parameterized counterpart of the curvature of the 2-D signer’s wrist trajectory to identify keyframes, have been recently reported in the literature. In this paper we extend these ideas by modeling the 3-D hand motion that is extracted from each frame of the video. To this end we propose a new informative function based on the $t$-parameterized curvature and torsion of the 3-D trajectory. The method to characterize video frames as keyframes depends on whether the motion occurs in 2-D or 3-D space. Specifically, in the case of 3-D motion we look for the maxima of the harmonic mean of the curvature and torsion of the target's trajectory; in the planar motion case we seek for the maxima of the trajectory's curvature.
The proposed 3-D feature is experimentally evaluated in applications of  sign language videos on (1) objective measures using ground-truth keyframe annotations, (2) human-based evaluation of understanding, and (3) gloss classification and the results obtained are promising.
\end{abstract}


\section{Introduction}\label{sec:intro}

The Sign Languages (SLs) are typically the main languages used by the  Deaf and by many of the hard-of-hearing (HoH).  Due to their poor experiences in spoken or written languages the Deaf almost always prefer SLs than reading or writing text \cite{nikolaraizi2013}. Nowadays, video capturing devices are ubiquitous and play an important role in the communication and education of the Deaf. A method to summarize SL videos, without sacrificing the semantics of the performed signs would offer significant benefits, especially in applications that require communication over low-bandwidth networks, or content browsing.

In the past, several general-purpose video summarization methods were presented (e.g., \cite{Rochan2018}, \cite{Guan2012}, \cite{Shao2009}); however those are not applicable in SL videos, since they treat the video frames holistically, while in the SL videos only some very specific regions are important for interpretation, while the rest of the frame is actually less relevant. Such regions are associated with specific parts of the human body, mainly the hands and face. This fact stands in stark contrast to holistic summarization methods.

The proposed method may have an impact on research related to the efficient transmission of SL videos over low-bandwidth networks. It is also expected to have indirect impact on the analysis of SL, by focusing on the frames where the most important information is present. This could result in significant savings in the recognition process, such as by using only a small portion of the data as input to a classifier.

%
Sign phonology includes the description of the handshape (hand formation), the movement, the location and the orientation (palm and fingers). In relation to the movement \cite{Crasborn2001} notes that the wrist moves through space in order to achieve a change of location. Furthermore, \cite{Pfau2012} refer to the sonority of syllables as the ability of a sign to be perceived at greater distance. Therefore, joints closer to the body are considered to be higher in the rank of sonority. \cite{Brentari1998} proposes a sonority hierarchy as follows: Shoulder - elbow - wrist - base joints - non-base joints. The above underline the importance of the wrist for the perceptions of the movement of a sign.

In this work, building upon the ideas presented in \cite{sartinas20212}, we contribute by
 introducing a method for efficient summarization of SL videos, based on wrist motion, that preserves their lexical meaning. Specifically, we use the $t$-parameterized Frennet-Serret frame to trace the signer's wrist and introduce a new informative function based on the curvature and torsion of the curve. These fundamental features measure how much the trajectory bends as it evolves and are used to identify the keyframes in the sign language video. 

The rest of the paper is structured as follows: Section \ref{sec:related} presents the prior work. Section \ref{sec:method} contains  the problem formulation while in Section \ref{sec:proposed} the proposed methodology is presented, which is followed by the  experimental results in Section \ref{sec:experiments}. Finally, Section \ref{sec:conclusion} concludes this paper.

\section{Related Work}\label{sec:related}

A related line of research deals with the extraction of keyframes from video content, also called ``static summary'' (in contrast to ``dynamic summary'' that extracts short videos). Many initial approaches used low-level features such as the color or motion histograms (e.g., \cite{Shao2009}), SIFT/SURF (e.g., \cite{Guan2012}), or more recently features from pretrained CNNs \cite{Rochan2018}. Then the keyframes are typically extracted using entropy (e.g., \cite{Cernekova2002}) or clustering methods (e.g., \cite{Chasanis2008}). Such methods mainly use the structural and not the semantic information in the video; however, they count on the fact that the changes in the structural frame data (objects) may be associated to semantic changes, which quite often is true. Some later approaches try to identify the semantic events, which are of importance, like in sports, e.g., \cite{Agyeman2019} or video surveillance, e.g., \cite{Lai2016}. To this end, objects may be identified and tracked.    

There have been reported supervised methods, which assume human annotations of keyframes in training videos, and seek to optimize the frame selection by minimizing loss with respect to this ground truth. In \cite{Zhang2016}, two LSTMs are used to select keyframes, by minimizing the cross-entropy loss on annotated ground-truth keyframes with an additional objective based on determinantal point process (DPP) to ensure diversity of the selected frames. In our method we don't optimize an objective using ground-truth labels, so our method belongs to the unsupervised ones.

Some of the most recent works in unsupervised summarization exploit the autoencoder architecture combined with recurrent networks such as the LSTM. In \cite{Yang2015} the auto-encoder is trained using a proposed shrinking exponential loss function that makes it robust to noise in the web-crawled training data, and is configured with bidirectional long-short-term memory (LSTM) cells to better model the temporal structure of highlight segments. In \cite{Mahasseni2017} the summarizer is the autoencoder long short-term memory network (LSTM) aimed at, first, selecting video frames, and then decoding the obtained summarization for reconstructing the input video.

Another line of research aims to find some trade-offs for the transmission of SL videos via low-bandwidth networks without sacrificing comprehensibility. Video coding systems use the particular structure of SL, i.e., the fact that the hands and face are the most important and thus need higher quality of representation, e.g., \cite{AGRAFIOTIS2006}, \cite{Saxe2002}. 
Of more relevance to our work is the laboratory study in \cite{Tran2014}, which identified the lower threshold for real-time conversations to be intelligible.
The study found that it was possible to hold a discussion even at a rate of 5fps, although the signers would need to sign at a slower speed. In the general case, a threshold of 10fps at 50kbps was found to be satisfactory.

The summarization of SL videos has only been addressed in a few studies. In \cite{Kosmopoulos2005} the region of hands and face are segmented using skin color and then modeled using Zernike moments. The second derivative of the moment norm may be used to extract the keyframes, assuming these are the turning points in the overall motion. While the method provides reasonable results, the shapes are often inaccurately represented and calculating higher order moments for better representation can be demanding. A related method is proposed in \cite{geetha2013dynamic}, where the keyframes for compressing sign videos and solving sign classification problems are identified as the frames corresponding to the Maximum Curvature Points (MCPs) of the global trajectory. 
However, this feature is invariant to the motion speed variability that may occur when different persons perform the same gesture making its use unsuitable for describing the dynamics of the motion model, which are critical for selecting the keyframes of a gloss.

The use of keyframes in translation and/or classification problems has been treated in few works related with sign language or action recognition. Specifically, in \cite{8846585}, a keyframe extraction method, called adaptive clip summarization where the proposed scheme automatically obtains variable-sized key frames/clips and implements dynamic temporal pooling on less-important frames/clips, was used. Then, the compact vectors are  considered as visemes/signemes under respective RGB/skeleton channels. 

In \cite{wang2018temporal}, the authors, motivated by the observation that consecutive frames are highly redundant, developed a video-level framework, called temporal segment network (TSN). This framework extracts short snippets over a long video sequence with a sparse sampling scheme, where the  samples are distributed uniformly along the temporal dimension. Finally, in \cite{zhu2016key}, the authors extract key volumes in both temporal and spatial domain, in order to improve the system's classification performance. The method identified key volumes simultaneously with classification.

In contrast to our work, all of these methods are applied to classification or translation tasks and require training for efficient summarization of SL videos.

\section{Problem Formulation}
\label{sec:method}

The inability of the human visual system to perceive and track fast motions has been extensively exploited to reduce the high computational cost in many applications in the field of computer graphics \cite{chalmers2003visual}. This is because 
the velocity of motion, which is a fundamental characteristic of movement in general, strongly affects the intelligibility of sign language.

It is common for videos to have at least one significant object in motion, often including the camera itself.
In a sign language video there are some frames, which offer clear hand shapes and others that suffer from motion blur. The blurry frames most often correspond to abrupt hand motion that the camera struggles to capture, but typically such frames are not semantically significant \cite{dilsizian2014}. In addition, not all the frames showing sharp handshapes are essential for the intelligibility of the sign, as they may simply repeat the same view. Based on this fact, we anticipate that certain keyframes within a sign are crucial for comprehending its meaning, while the majority of video frames are not critical at all. This is a rule, rather than an exception in SLs, that we are going to exploit to extract summaries of SL videos while preserving their intelligibility.

\begin{figure}[!t]
\centering
\includegraphics[width=6in]{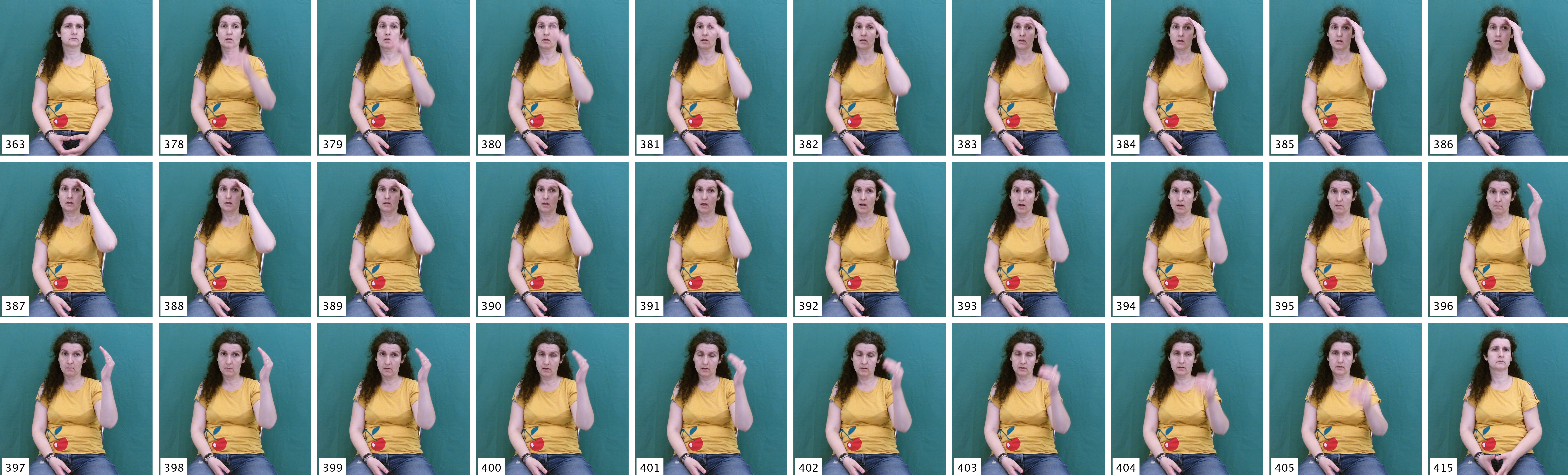}
\caption{The consecutive frames from rest to rest position corresponding to the sign  ``$\kappa\alpha\lambda\eta\mu\acute{\epsilon}\rho\alpha$'' (good morning)}
\label{fig:montage}
\end{figure}

In order to identify the keyframes of a SL video we are going to introduce a new feature function, which is based on the harmonic mean of the $t$-parameterized curvature and torsion  of the trace of the signer's wrist. To this end, let us define the following pointset:

\begin{equation}
\mathbb{P}_N=\left\{\tilde{\mathbf{r}}_n\right\}^N_{n=1}  \label{pointset}
\end{equation}

with $\tilde{\mathbf{r}}_n~\in \mathbb{R}^3$ denoting the coordinates of the signer's wrist at the $n-th$ frame of the video with respect to the frame's upper left corner. Note that the aforementioned pointset, denotes the trace of the signer's wrist in 3-D Euclidean space $ \mathbb{R}^3$, as it  moves from its starting (rest) position to its ending one as shown for the sign in  Fig.\ref{fig:montage}a.
In order to make mathematics tractable,  let us consider that the elements of the pointset $\mathbb{P}_N$ result from sampling the following 3-D differentiable curve:

\begin{equation}
\tilde{\mathbf{r}}(t)=[\tilde{x}(t)~\tilde{y}(t)~\tilde{z}(t)]^T,~t\in (t_0,~T) \label{t_trajectory}
\end{equation}

that represents the time - parameterized trajectory of a particle as it moves in the 3-D Euclidean space $\mathbb{R}^3$ for the time interval $[t_0,~T]$.
We assume that the velocity vector $\tilde{\mathbf{r}}^{(1)}(t)$, the acceleration vector $\tilde{\mathbf{r}}^{(2)}(t)$ and the jerk vector $\tilde{\mathbf{r}}^{(3)}(t)$ are not proportional to each other. Let $s(t)$ denote the arc length covered by a particle along the curve $\tilde{\mathbf{r}}(t)$ over the time interval $[t_0, ~t]$, that is:

\begin{equation}
s(t)=\int_{t_0}^t ||\tilde{\mathbf{r}}^{(1)}_{t}(\tau)||_2d\tau, \label{arc_length}
\end{equation}

with $||\mathbf{p}(.)||_2$ and $\mathbf{p}^{(1)}_{w}(.)$ denoting the $l_2$ norm of 3-D vector function $\mathbf{p}(.)$ and the derivative of the function $\mathbf{p}()$ with respect to the variable $w$ respectively. 

The 3-D curve defined in Eq. ({\ref{t_trajectory}}) can be re-parameterized by its arc length $s$ as:

\begin{equation}
\mathbf{r}(s)=[x(s)~y(s)~z(s)]^T,~s\in[s(t_0),~s(T)]. \label{s_trajectory}
\end{equation}

Note that since the function defined in Eq. (\ref{arc_length}) is an increasing function of $t$, its inverse function, let us denote it by $t(s)$, exists and is a well-defined function. Based on that fact, the $t$ and $s$ parameterized trajectories, although they are distinct functions, are related by the following equations: $\tilde{\mathbf{r}}(t) = \mathbf{r}(s(t))$ and $\mathbf{r}(s) = \tilde{\mathbf{r}}(t(s))$.
Note also that by parameterizing the trajectory of the particle by the arc length, its description does not depend on the rate quantified by the time derivative of the arc length, i.e. $s^{(1)}(t)$, in which the particle has traversed it. Proposition 2 proves that the number of $t$-parameterized  trajectories corresponding to the same $s$-parameterized one defined in Eq. (\ref{s_trajectory}) is infinite.

\subsection{Frenet-Serret Frame} 
Using  the $s$-parameterized trajectory defined in Eq. (\ref{s_trajectory}), Frenet-Serret $\mathbf{tnb}$ frame is defined through the following three orthogonal vectors:
\begin{equation}
\mathbf{t}(s)=\frac{\mathbf{r}^{(1)}_{s}(s)}{||\mathbf{r}^{(1)}_{s}(s)||_2},~ \mathbf{n}(s)=\frac{\mathbf{t}^{(1)}_{s}(s)}{||\mathbf{t}^{(1)}_{s}(s)||_2},~
\mathbf{b}(s)= \mathbf{t}(s) \times \mathbf{n}(s) \label{eq:tnb}
\end{equation}

where $\mathbf{f}^{(1)}_{s}(s)$,  ``$\times$''  denote the derivative of the vector function  with respect to arc length $s$ of the 3-D curve and the cross-product operator respectively. More precisely, 
the tangent vector  $\mathbf{t}(s)$ points to the direction the curve travels, the normal vector $\mathbf{n}(s)$ is orthogonal to the tangent, while binormal vector $\mathbf{b}(s)$ is orthogonal to the plane defined by the aforementioned pair of orthonormal vectors, as it can be validated by Eq. (\ref{eq:tnb}). Note also that any pair of the above defined vectors can be used for defining a specific plane; namely:
\begin{itemize}
    \item the osculating $\mathbf{t-n }$,
    \item the normal  $\mathbf{n-b }$ and
    \item the rectifying $\mathbf{b-t}$ respectively.
\end{itemize}    
Based on the orthogonality of the above mentioned vectors, a local, data-dependent orthonormal basis for the space $\mathbb{R}^3$ is derived. 

By defining now the matrix function:
\begin{equation}
R(s)=[ \mathbf{t}(s)~\mathbf{n}(s)~\mathbf{b}(s)]\label{matrix_function}
\end{equation}

whose columns contain the above defined unit vector functions, the $\mathbf{Frenet-Serret}$ frame can be defined by the following differential equation:
\begin{equation}
R^{(1)}_{s}(s)=R(s)C(s) \label{FS_matrix}
\end{equation}

where $X^{(1)}_{s}(s)$, denotes the derivative of the matrix function $X(s)$ with respect to arc length  $s$, and matrix  $C(s)$ an antisymmetric matrix which is defined as follows:
\begin{equation}
C(s)=\left[\begin{array}{ccc}  0 & -\kappa(s) & 0 \\
\kappa(s) & 0 & -\tau(s) \\
0 & \tau(s) & 0 \end{array} \right ] 
\end{equation}

with $\kappa(s)$, $\tau(s)$ denoting the \textbf{curvature} and the \textbf{torsion} of the curve respectively.  

Note that Frenet-Serret frame can be defined and used for the description of both 2-D and 3-D curves that are not straight lines; for straight lines the curvature is equal to zero and the torsion is undefined. Moreover, in the case of planar (that is 2-D) curves, the torsion is equal to zero; the curvature at a point of such a differentiable curve is defined as the reciprocal of the radius of its osculating circle, that is the circle that best approximates the curve near that point. It is clear that the smaller this circle is the higher its curvature, with its units being $m^{-1}$ in International System of Units (SI).

For the 3-D curves, the torsion measures how sharply the curve is twisting out of the plane of curvature, i.e. the osculating $\mathbf{t-n}$ plane. It has the same units with the curvature and taken together can be used for the perfect reconstruction of a 3-D  curve.

In order to take into account the existing particularities in the description of the 2-D and 3-D curves and define an appropriate measure that uses in a natural manner the above mentioned quantities, using $\mathbf{tnb}$ frame,  we are going to prove the following proposition.

\textbf{Proposition 1: } Let $\mathbf{r}_{s}^{(n)}(s),~n=1,~2,~3,~||\mathbf{r}^{(1)}_{t}(s(t))||_2$ be the $n-$th order derivative of the $s$-parameterized trajectory of the curve defined in Eq. (\ref{s_trajectory}) with respect to its arc length $s$ (that is the $s$-parameterized counterparts of the velocity, acceleration and the jerk) and the speed $v(t)$ respectively. Then, the following $s$-parameterized based relations hold: 

\begin{align}
\mathbf{r}^{(1)}_{s}(s)~\times~\mathbf{r}_{s}^{(2)}(s)&=\kappa(s)\mathbf{b}(s)\label{curv}\\
\left\langle\mathbf{r}^{(1)}_{s}(s)~\times~\mathbf{r}^{(2)}_{s}(s),~\mathbf{r}^{(3)}_{s}(s)\right\rangle&=\tau(s)||\mathbf{r}^{(1)}_{s}(s)~\times~\mathbf{r}^{(2)}_{s}(s)||_2^2, \label{tor}
\end{align}
where $\langle\mathbf{x},\mathbf{y}\rangle$ denotes the inner product of the vectors $\mathbf{x},~\mathbf{y}$, and $\kappa(s),~\tau(s)$ are the  curvature  and  torsion respectively. 

\textbf{Proof:}
The proof of the proposition goes as follows.  
Let $\mathbf{r}_{t}^{(1)}\big(s(t)\big)$ be the derivative of the function  $\mathbf{r}\big(s(t)\big)$ (that is the composition of functions $\mathbf{r}(s)$ and $s(t)$) with respect to time variable $t$. Then, using the chain rule the following relation holds:
\begin{equation}
\mathbf{r}^{(1)}_{t}\big(s(t)\big)=\frac{d\mathbf{r}\big(s(t)\big)}{dt}=\frac{d\mathbf{r}\big(s(t)\big)}{ds(t)}\frac{ds(t)}{dt}=\mathbf{r}^{(1)}_{s}\big(s(t)\big)v(t)
\end{equation}

where $\mathbf{r}\big(s(t)\big)$, $v(t)$ denote the $t$-parameterized counterpart of $\mathbf{r}(s)$ and the speed of the moved particle respectively. Thus, given that 
$||\mathbf{r}^{(1)}_s(s)||_2=1$ (since $||\mathbf{r}^{(1)}_{t}\big(s(t)\big)||_2=v(t)$), the first component of the Frenet-Serret frame of  Eq. (\ref{eq:tnb}) can be rewritten as:
\begin{equation}
\mathbf{r}^{(1)}_s(s)=\mathbf{t}(s).\label{r1}
\end{equation} 

By differentiating both sides of this relation with respect to the arc length and using the Frennet-Serret frame defined in Eq. (\ref{FS_matrix}), the $s$-parameterized counterpart of the vector acceleration function can be defined as follows:
\begin{equation}
\mathbf{r}^{(2)}_s(s)=\kappa(s)\mathbf{n}(s). \label{r2}
\end{equation}

Note that this function is collinear with the normal vector $\mathbf{n}(s)$ of the Frenet-Serret frame. Using this relation, Eq. (\ref{r1}) and the definition of the cross product, the relation (\ref{curv}) of the proposition can be easily obtained.

Let us prove the relation (\ref{tor}) of the proposition. In order to achieve our goal we differentiate both sides of Eq. (\ref{r2}) with respect to the arc length $s$ and use again the Frennet-Serret frame defined in Eq. (\ref{FS_matrix}) to obtain:
\begin{equation}
\mathbf{r}^{(3)}_s(s)=-\kappa^2(s) \mathbf{t}(s)+\kappa^{(1)}_s(s)\mathbf{n}(s)+\kappa(s)\tau(s)\mathbf{b}(s)\label{r3}
\end{equation}

where $\kappa^{(1)}_s(s)$ denotes the first order derivative of the curvature with respect to the arc length $s$. Note that the $s$-parameterized counterpart of the jerk function belongs in the 3-D space, i.e., $\mathbb{R}^3$. Using this relation, Eq. (\ref{curv}), the orthonormality of the Frennet-Serret frame and the definition of the dot product, the relation (\ref{tor}) of the proposition can be easily obtained.
\hfill $\square$

 Using Proposition 1, we can define the curvature as well as the torsion by the following relations: 

\begin{eqnarray}
\kappa(s)&=&||\mathbf{r}^{(1)}_s(s)~\times~\mathbf{r}^{(2)}_s(s)||_2 \label{k(s)}\\
|\tau(s)|&=&\frac{\left|\left\langle\mathbf{r}^{(1)}_s(s)~\times~\mathbf{r}^{(2)}_s(s),~\mathbf{r}^{(3)}_s(s)\right\rangle\right|}{\kappa^2(s)} \label{t(s)}
\end{eqnarray}
where $|x|$ denotes the absolute value of the scalar quantity $x$. 

It is clear from Eq.(\ref{t(s)}), that torsion can be defined only when curvature is strictly positive  that is,  when  the particle's trajectory does not coincide to a straight line. From this point on we consider that it holds. 
Note, as it was aforementioned, that the units of the quantities  defined in Eqs. (\ref{k(s)}) and (\ref{t(s)}) are the same; these quantities  are strongly related to the binormal vector $\mathbf{b}(s)$. The $s$-parameterized curvature $\kappa(s)$ and torsion $\tau(s)$, as seen from Eqs. (\ref{curv}) and (\ref{tor}), define two different instantaneous radii for the composite total motion. 

Let us concentrate ourselves on the basic drawback of the above defined $s$-parameterized quantities. 
It is worth noting that the $s$-parameterized curvature and torsion remain unchanged despite the particle's motion, as the same displacement can be achieved by the particle moving with a constant velocity, also known as the \textbf{average velocity}, within the same time interval.
This means that the $s$-parameterized curvature and torsion can be considered as static or shape descriptors, which do not describe the \textbf{kinematic model} of the particle, and are thus unsuitable for it.
Indeed, as it was pointed out in \cite{sartinas20212} through the simple but informative case of the \textbf{circular motion} the $s$-parameterized curvature, that is $\kappa(s)=r^{-1}$, does not depend on the motion model of the particle. 

In the following proposition we prove rigorously a more general result regarding the $s$ parameterization based 3-D curves, that definitely points to the aforementioned direction.





\textbf{Proposition 2: }  Let $\tilde{\mathbf{r}}_1(t)$ be a $t$-parameterized trajectory  of a particle that is moving into $\mathbb{R}^3$ for the time interval $[t_0,~T]$ and $f(t)$ a continuous and increasing function of $t$ in the same time interval with its initial and final value satisfying $f(t_0)=t_0$ and   $f(T)=T$. Let us also consider that   $\tilde{\mathbf{r}}_2(t)=\tilde{\mathbf{r}}_1(f(t))$  is the $t$-parameterized  trajectory of the particle resulting from the composition of the  $\tilde{\mathbf{r}}(t)$ with function $f(\cdot)$. 
Then, their $s$-parameterized counterparts coincide, i.e., $\mathbf{r}_2(s)=\mathbf{r}_1(s)$.


\textbf{Proof:}
For the archlength $s_2(t)$ of the second trajectory, we have:
\begin{align}\label{eq:integration}
s_2(t) &= \int_{t_0}^t ||\tilde{\mathbf{r}}_{2~\tau}^{(1)}(\tau)||_2d\tau \nonumber\\
&=\int_{f(t_0)}^{f(t)}||\tilde{\mathbf{r}}_{1~f(\tau)}^{(1)}\Big(f(\tau)\Big)||_2df(\tau),
\end{align}
and since $f(t_0)=t_0$
the following relation holds:
\begin{equation}
s_2(t)=s_1(f(t)).\label{equiv}
\end{equation}

Let us define now the following inverse functions:
\begin{equation}
t_i(s)=s_i^{-1}(s),~i=1,2.\nonumber
\end{equation}

Based on the monotonicity of the functions $s_i(t),~i=1,2$ as well as of the function $f(t)$ the above defined inverse functions are well defined. Using the definition of the inverse function $t_1(s)$ and Eq. (\ref{equiv}) the inverse function $t_2(s)$  can be expressed as follows:
\begin{equation}
t_2(s)=s_2^{-1}(s)=f^{-1}\big(s_1^{-1}(s)\big)\label{eq:inv_traj2}.
\end{equation}

The $s$-parameterized trajectories of the particles can now easily obtained. Indeed,  for the first particle we have:
\begin{equation}\label{eq:r_1s}
    \mathbf{r}_1(s) = \tilde{\mathbf{r}}_1\big(t_1(s)\big) = \tilde{\mathbf{r}}_1\big(s_1^{-1}(s)\big),
\end{equation}

while for the second one, using Eq. (\ref{eq:inv_traj2}):
\begin{align}\label{eq:r_2s}
    \mathbf{r}_2(s) &= \tilde{\mathbf{r}}_1\Big(f\big(t_2(s)\big)\Big)\nonumber\\
&=\tilde{\mathbf{r}}_1\Big(f\big(f^{-1}(s_1^{-1}(s))\big)\Big)=\tilde{\mathbf{r}}_1\big(s_1^{-1}(s)\big).
\end{align}
From Eqs. (\ref{eq:r_1s}) and (\ref{eq:r_2s}) we have that $\mathbf{r}_2(s)=\mathbf{r}_1(s)$ and this  concludes the proof of the proposition.
\hfill $\square$


\section{The Proposed Solution }
\label{sec:proposed}
To overcome the limitations of the $s$-parameterized curvature and torsion in representing the kinematic model we propose the use of their $t$-parameterized counterparts. 
Let us define the $t$-parameterized counterparts of the Frenet-Serret $\mathbf{tnb}$ frame  defined in Eq. (\ref{eq:tnb}) by substituting the arc length variable $s$ with the function defined in Eq. (\ref{arc_length}) and defining the following matrix function:
\begin{equation}
\tilde{R}(t)=[ \mathbf{t}\big(s(t)\big)~\mathbf{n}\big(s(t)\big)~\mathbf{b}\big(s(t)\big)]\label{t-matrix_function}
\end{equation}

which constitutes the $t$-counterpart of the matrix function defined in Eq. (\ref{matrix_function}).
Then, the $t$ Frenet-Serret $\mathbf{tnb}$ frame can be defined by the following differential equation:
\begin{equation}
    R^{(1)}_{t}(t)=\tilde{R}(t)\tilde{C}(t) \label{t-FS_matrix}
\end{equation}

and the elements of the matrix $\tilde{C}(t)$ that constitute the $t$-parameterized counterparts of the curvature and torsion, are defined as follows:
\begin{eqnarray}
K(t)&=&\kappa\big(s(t)\big)v(t) \label{k(t)}\\
T(t)&=&\tau\big(s(t)\big)v(t) \label{t(t)}
\end{eqnarray}
with $v(t)$ denoting the speed of the particle.

It is clear from Eqs. (\ref{k(t)}) and (\ref{t(t)}) that the units of the proposed re-parameterized curvature and torsion are $sec^{-1}$, i.e., Hz and consequently they can be considered as the \textbf{instantaneous ordinary frequencies} of the composite motion of the particle as it moves in the 3-D space.
As pointed out in \cite{sartinas20212}, the proposed descriptor not only represents the related \textbf{shape} information of the trajectory, but also the \textbf{kinetics}, i.e., the dynamics, of the motion. This is exemplified in the case of planar simple circular motion where $K(t)=\theta^{(1)}(t)$, where the phase function $\theta(t)$ describes the motion model of the particle.


Moreover, adopting the proposition of\cite{sartinas20212}, the frames in SL videos that contain the highest $t$-parameterized curvature, denoted by $K(t)$, can be used as keyframes to achieve highly compressed videos while maintaining their intelligibility.



We must stress at this point that the maxima of the $s$-parameterized curvature $|\kappa(s)|$ was proposed for identifying keyframes in \cite{geetha2013dynamic} and this concludes our analysis for the planar (that is 2-D) curves.


\subsection{3-D Curves - Harmonic Mean of Curvature and Torsion}

We propose using the harmonic mean of curvature and torsion as a figure of merit for selecting the most informative frames in 3-D curves. The harmonic mean is appropriate for situations when the \textbf{average of rates} is desired \cite{ferger1931nature}. Since the $t$-parameterized quantities represent instantaneous frequencies closely related to rates, using their \textbf{harmonic mean} is a reasonable choice. In 3-D motion, the torsion $T(t)$ is not zero, and the harmonic mean is defined as:

\begin{equation}
   H(t)= \displaystyle{\frac{2K(t)|T(t)|}{K(t)+|T(t)|}}\label{hm}
\end{equation}

However, when the 3-D motion degenerates to a planar one the $t$-parameterized Curvature proposed in \cite{sartinas20212} is adopted, so we have the following cases:
\begin{equation}
M(t)=
\begin{cases}
H(t), & \text{ in the case of 3-D motion}\\
K(t), & \text{in the case of 2-D motion.}
\end{cases}\label{metric_M}
\end{equation}

We seek for maxima of this \textbf{metric}, i.e.:

\begin{equation}
M^{(1)}(t)=0 ~~\text{and} ~M^{(2)}(t)<0.\label{harmonic}
\end{equation}

The $H(t)$ expresses the reciprocal of the arithmetic mean of the instantaneous radiuses where the ``mean'' motion should occur.

Note that in order to chose the most appropriate metric in Eq. (\ref{metric_M}), first we have to solve a classification problem regarding to the kind of the particle's motion and in particular the dimensionality of its motion. As we are going to see in the next section, in order to achieve our goal, we adopt a PCA-based approach in each rest-to-rest (R2R) signing interval.

\section{Experimental results}
\label{sec:experiments}

This section will present the results obtained by applying our proposed figure of merit as well as other state-of-the-art techniques \cite{geetha2013dynamic, Kosmopoulos2005, sartinas20212} to the only available annotated dataset with keyframes that we are aware of \cite{sartinas20212}.

We will follow an experimental procedure similar to the one described in \cite{sartinas20212}. It is clear that the first we have to do for the application of the proposed technique is to detect the appropriate branch in Eq. (\ref{metric_M}) by classifying the wrist motion in each signing interval. The hand pose is always measured in 3D, but its motion can still be planar, or close to planar depending on the uttered sign. In order to classify a R2R signing interval as planar or non-planar the 3-D wrist trajectory of the signing interval is fitted with a plane using PCA. To this end if $\mathcal{S}=\{\mathbf{r}_n\}_{n=1}^N$ is the set of trajectory points in the  R2R signing interval, we form their $3\times 3$ covariance matrix and the fitting error is calculated from the ratio $\sigma_3 / (\sigma_1 + \sigma_2+\sigma_3)$, with $\sigma_i$ denoting the $i$-th singular value of the aforementioned matrix. If the resulting \textbf{fitting error} is smaller than a predefined value $f_{error}$, then the trajectory is considered planar and thus the second branch of Eq. (\ref{metric_M}) is used with the coordinates of the wrist points projected to the plane defined by the corresponding first two principal eigenvectors.

\begin{figure*}[!t]
\centering
\includegraphics[width=6in]{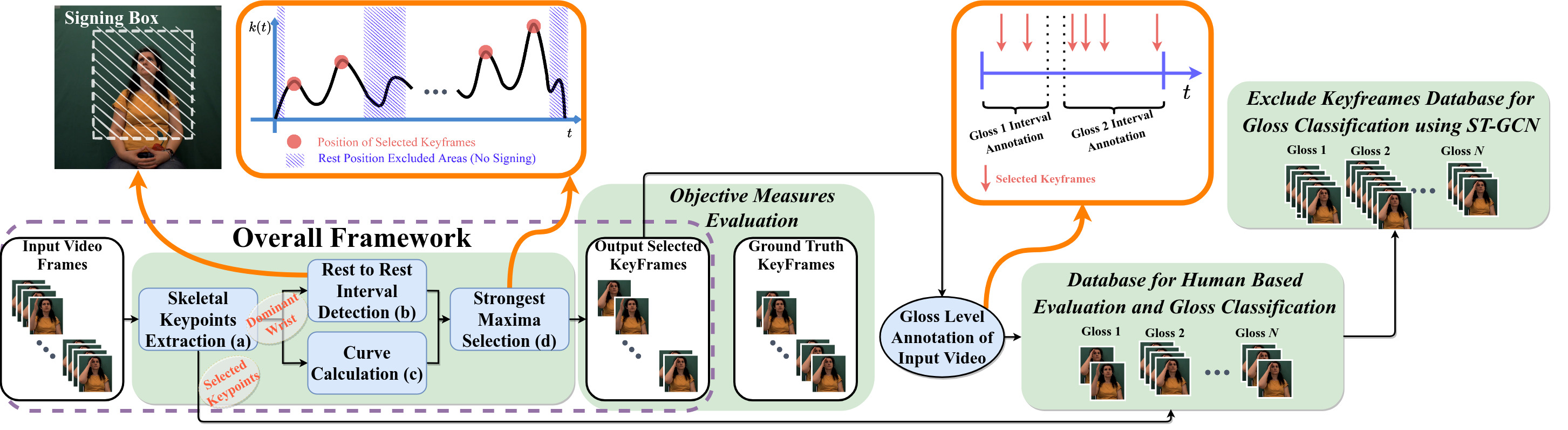}
\caption{The proposed overall summarization framework (dotted box). Its output is used for the objective measures evaluation (Section \ref{subsec:Objective}), by comparing it with the ground truth keyframes, and the creation of the database for human based evaluation (Section \ref{subsec:HumanBased}) and gloss  classification (Section \ref{subsec:GlossClassification})}
\label{fig:scheme}

\end{figure*}

\subsection{Experimental Setup}

We evaluated our method in a sign language dataset, which is manually annotated with keyframes. It was presented for the first time in \cite{sartinas20212}, and is composed by 32 videos of Greek SL of a total duration of 168 minutes containing approximately 5500 signs and a vocabulary of 387 unique glosses. Eight native signers performed four different scripts and were captured from a Ximea camera with 60 fps. 
The dataset was annotated by four experts in Greek SL who selected the least amount of keyframes for each gloss in order to be fully understandable. The range of the least amount of keyframes for each gloss was in the interval $[1, ~10]$. 

To assess the performance of the techniques, we compared them in (a) objective measures using the ground-truth, (b)  human-based evaluation of understanding and (c) gloss classification.

We compared our technique that is based on the metric $M(t)$, against the 2-D $s$-parameterized curvature $\kappa_{2D}(s)$ \cite{geetha2013dynamic}, the 3-D $s$-parameterized curvature $\kappa_{3D}(s)$ \cite{geetha2013dynamic}, the Zernike moment - based \cite{Kosmopoulos2005}, the 2-D $t$-parameterized curvature $K_{2D}(t)$ \cite{sartinas20212} and 3-D $t$-parameterized curvature $K_{3D}(t)$ that constitutes the 3-D extension of its 2-D counterpart.

In Fig. \ref{fig:scheme}  the overall proposed framework of summarizing SL video is depicted, that works for every video frame as follows: the skeleton tracker extracts among other keypoints the Signer’s dominant wrist position (a) which is used in turn for the detection of the signing intervals (b) and the calculation of the curve describing the importance of every frame (c). Finally, the important frames are selected using the positions, inside the signing intervals, where the selection curve attains its strongest maxima (d).

For the moments-based technique, we directly used the extrema of the produced curve instead of its second derivative proposed by the authors \cite{Kosmopoulos2005}, because it performed better in our dataset. The 3-D pointset of the signer's wrist trajectory was determined from MediaPipe skeleton tracker \cite{Lugaresi2019} and was smoothed by a Gaussian kernel  to eliminate the noise influence occurred from the wrist detection algorithm.  Subsequently, the informative points were identified, by applying the  methods  on the dataset. Since every technique leads to a different curve, whose number of extrema points may differ,  we used the prominence \cite{Fink} of every critical point as a measure of its importance. Hence, for a given video for all techniques, the desired number of keyframes were extracted  by selecting the strongest extrema based on their prominence values.  Finally, in all experiments we have conducted the value of the fitting error $f_{error}$ was set to 5e-2.

\subsection{Objective Measures Evaluation}
\label{subsec:Objective}

We assessed the methods performance by comparing the keyframes that they selected  with the ground truth one as depicted in Fig. \ref{fig:scheme}, using the well-known \textit{Recall} rate and $F_2$ score. In order to transform the problem at hand into a binary one, 
we consider that for each identified keyframe all neighbor frames that are within a temporal distance less than or equal to an experimentally defined threshold $\Delta$ are labeled by $1$. Following such a procedure we obtain a binary label for every video frame  we used for the evaluation. Because of the occurrence of many abrupt motions in SL, the threshold $\Delta$ in this experiment was set to 5 frames, which corresponds to 1/12 seconds since the camera frame rate was 60. 

\begin{figure*}[!t]
\centering
\subfloat[]{\includegraphics[width=1.3in]{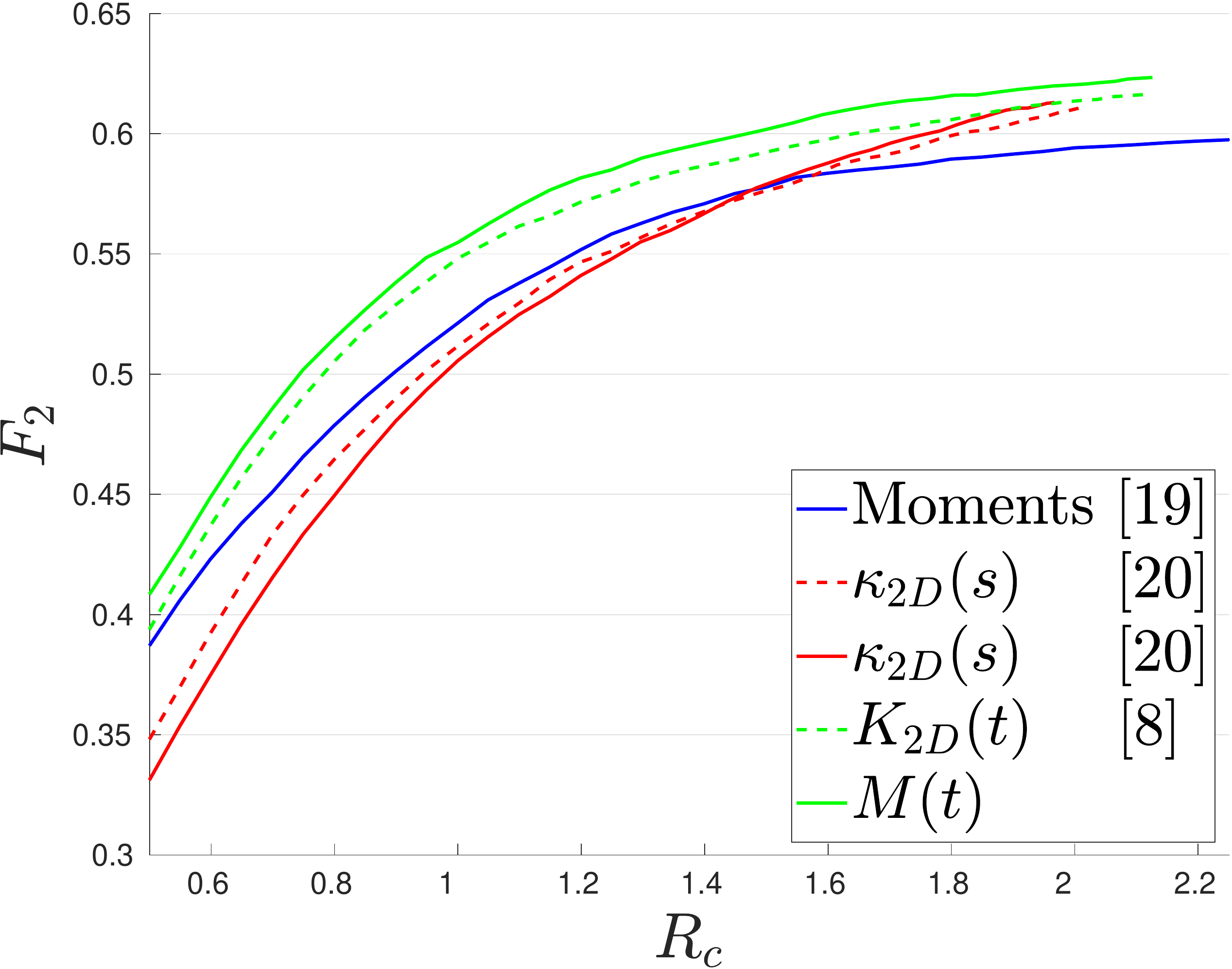}}
\subfloat[]{\includegraphics[width=1.3in]{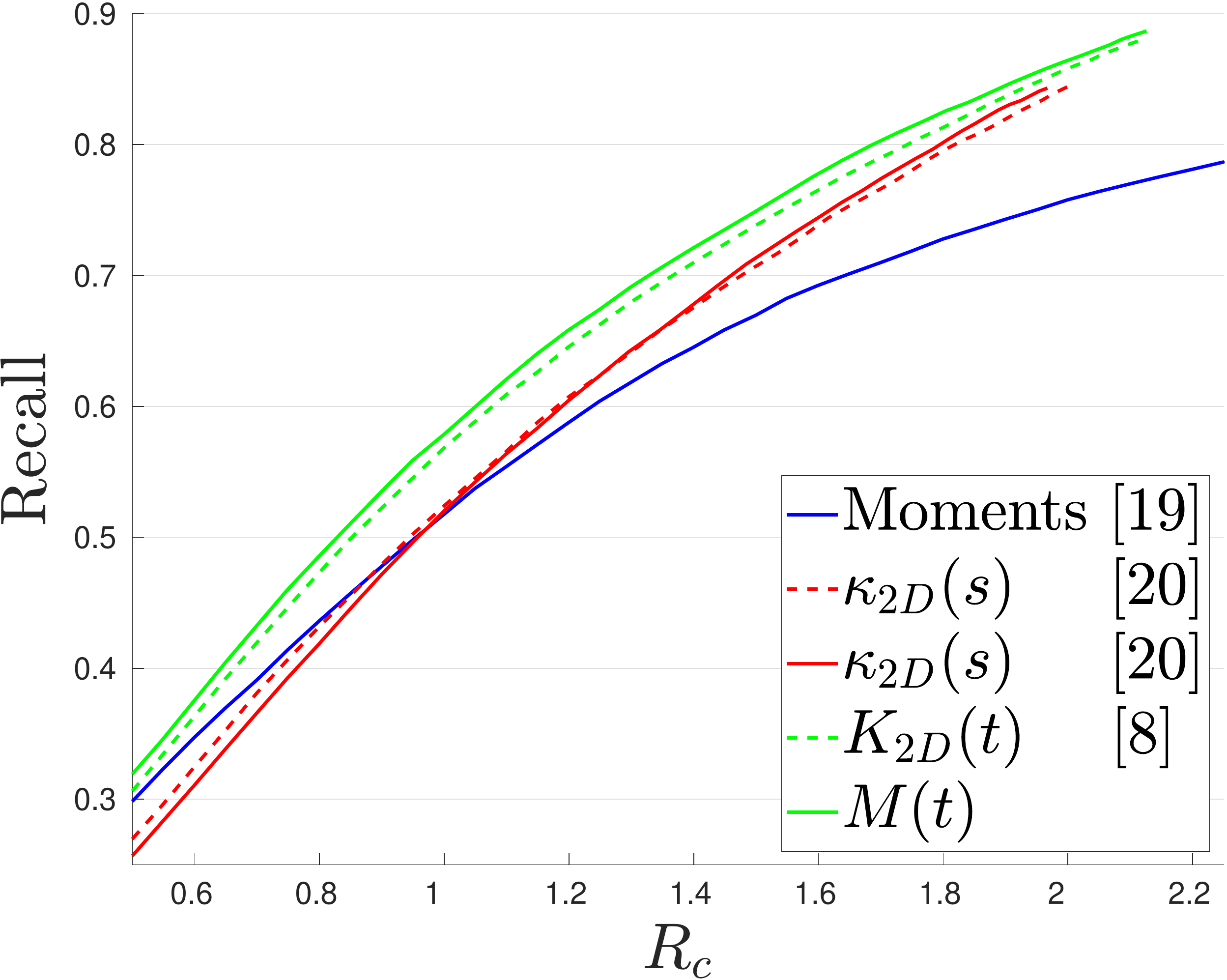}}
\subfloat[]{\includegraphics[width=1.3in]{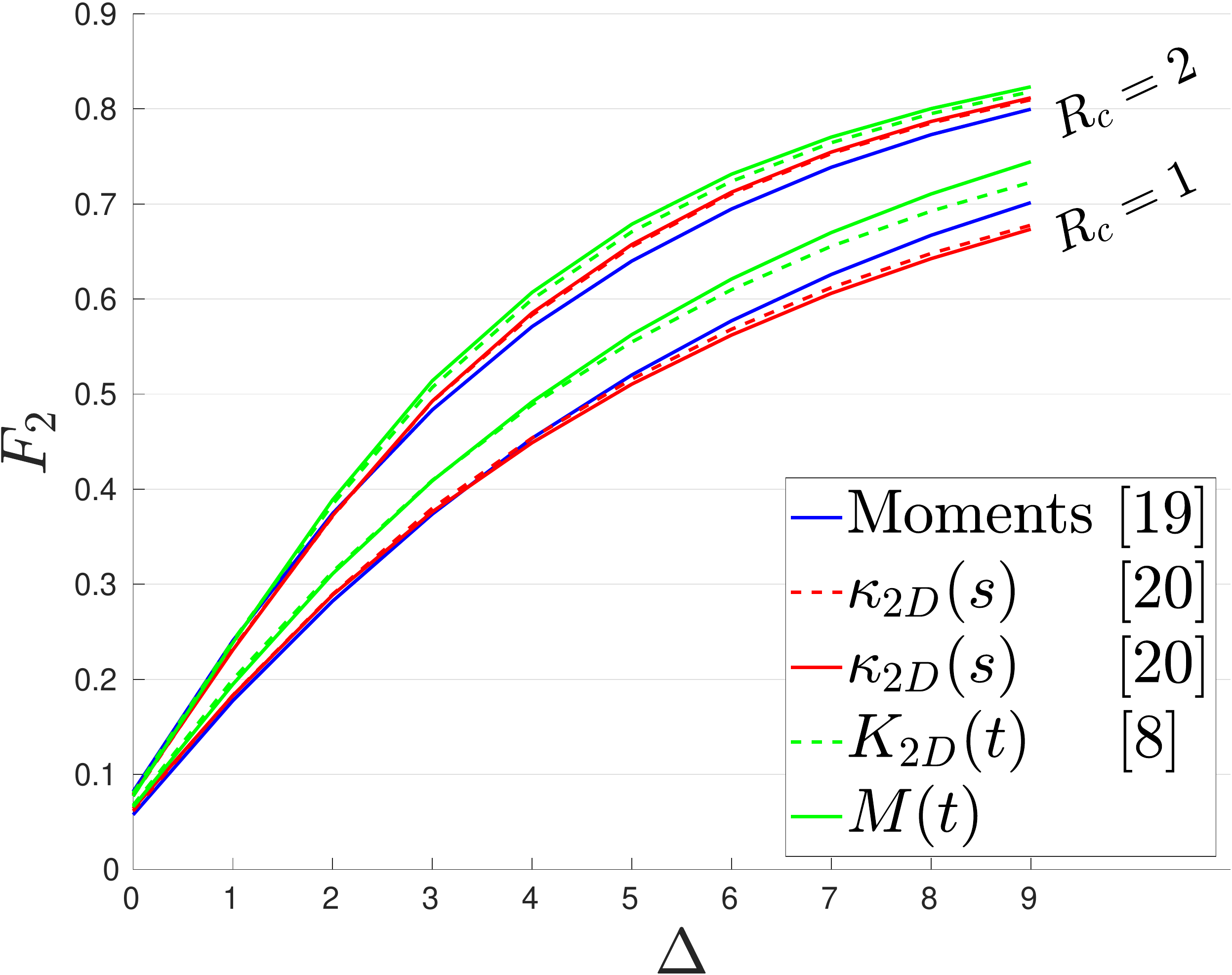}}
\subfloat[]{\includegraphics[width=1.3in]{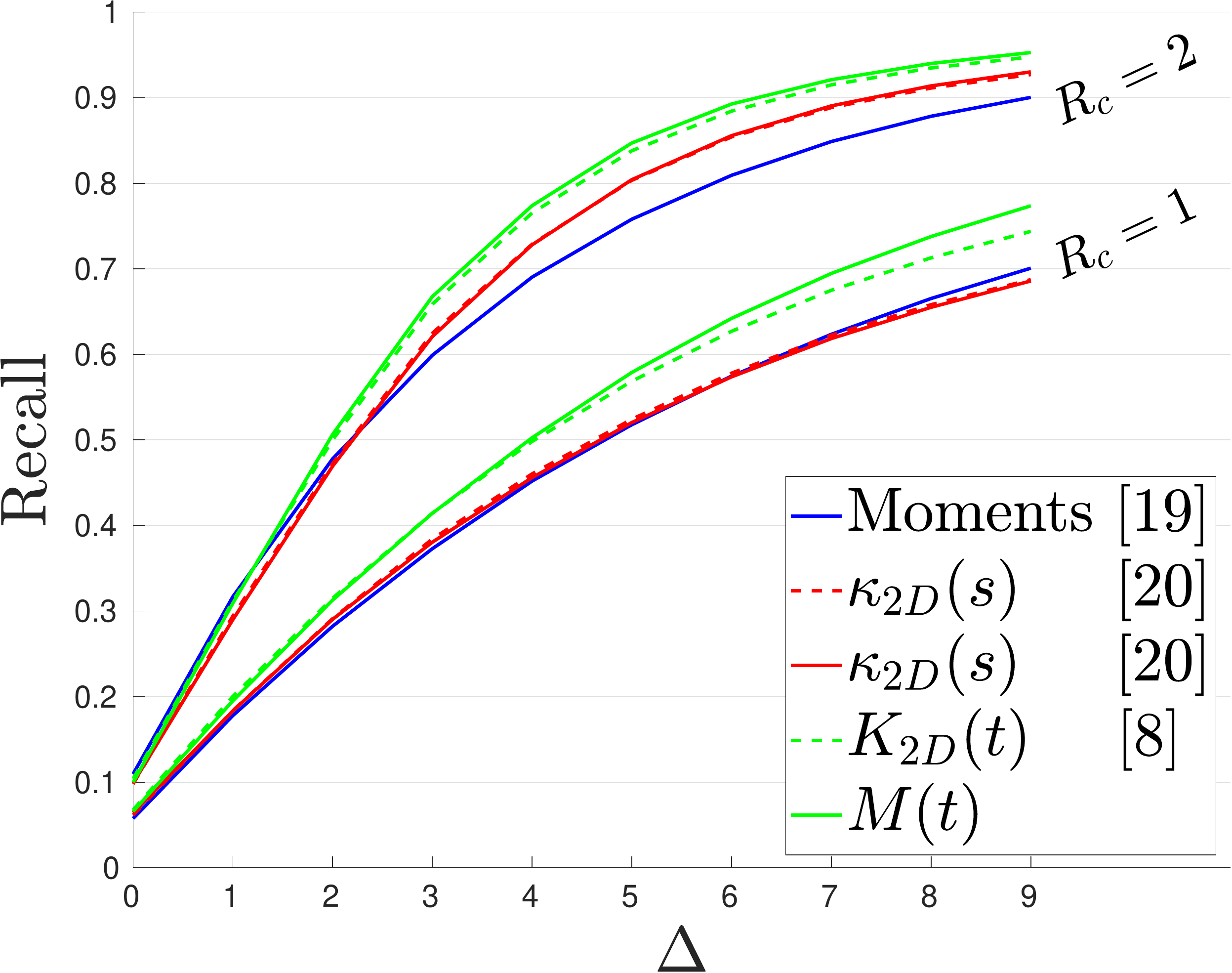}}
\subfloat[]{\includegraphics[width=1.3in]{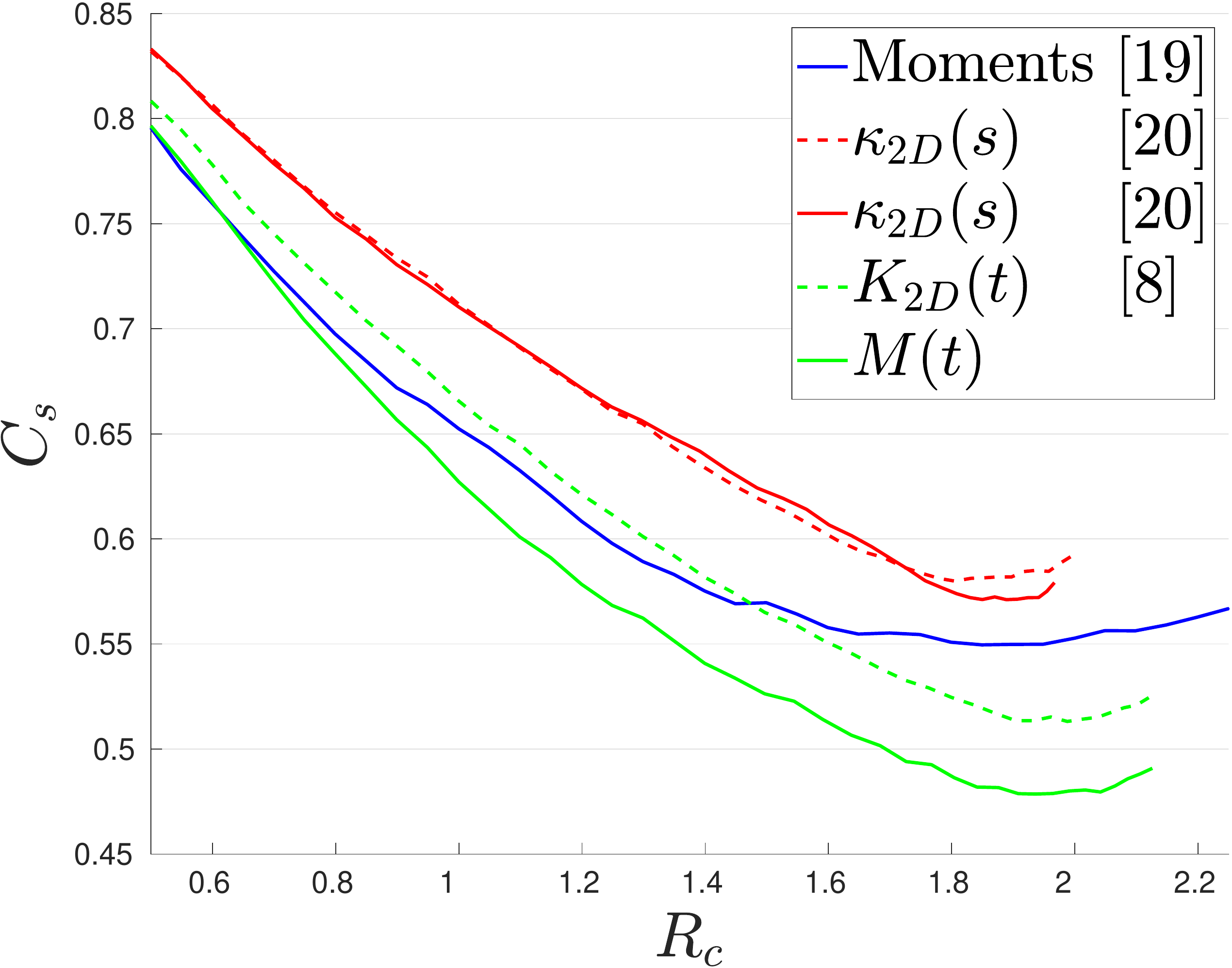}}
\caption{Obtained results in terms of (a, c) $F_2$ score, (b, d) \textit{Recall} rate and (e) relative mean captured sign's complexity metric $C_s$. Versus $R_c$ for $\Delta=5$ (a - b) and versus temporal proximity threshold $\Delta$ for $R_c=1, 2$ (c - d)}
\label{fig:delta_recall_f2}
\end{figure*}

The obtained results, in terms of recall and the $F_2$ score versus the ratio $R_c$, that is the number of selected keyframes by each technique to the number of the ones selected by the annotators, are shown in Fig. \ref{fig:delta_recall_f2}(a) and (b) respectively. It is evident that the proposed figure of merit outperformed the other methods. This in turn means that the proposed technique was in closer $\Delta$ proximity to the ground-truth and thus captured the meaning more accurately. Indeed, when for example the ratio $R_c=1$ the performance of the proposed  $t$-parameterized curvature in terms of the $F_2$ score was 4\% better than $s$-parameterized one and 3\% better than the moment-based technique. By taking into account the complexity of the problem at hand, we must stress at this point that the achieved $F_2$ score by the proposed criterion is very promising. 

Regarding the \textit{Recall} rate, comparative results are depicted in Fig. \ref{fig:delta_recall_f2}(b). The proposed technique achieved again better \textit{Recall} rate than the other ones. Indeed, the proposed technique selected the keyframes in such a way that 57\% of them were in $\Delta$ proximity with the corresponding ones of the annotators; for the $s$-parameterized curvature and moment-based technique the corresponding number was 52\%. 

Finally, in order to quantify the impact of the temporal threshold $\Delta$, we measured for different values of $\Delta$ the $F_2$ score and \textit{Recall} rates for $R_c=1$ and $R_c = 2$ (i.e., the amount of keyframes was set to be equal or twice as much as the ground truth one). The results are shown in Fig. \ref{fig:delta_recall_f2}(c) and (d). Clearly, both $F_2$ score and \textit{Recall} rate, are increasing functions of the proximity factor $\Delta$, retaining  methods ordering as well. In addition, as it was expected both metrics are increasing as the ratio $R_c$ increases.

 Note also that for both values of $R_c$ and for both objective evaluation metrics the proposed technique outperforms the other ones. This is more evident in the \textit{Recall} rate metric, that can be viewed as the probability that a relevant keyframe is selected by the technique and whose significance over \textit{Precision}, that can be considered as the probability that a keyframe randomly selected from the pool of total selected keyframes is relevant, has been indicated from the use of $F_2$ score instead of $F_1$ in our evaluation.

We also assessed the ability of the summarization methods to capture the complexity of signs  by extracting a  reasonable number of keyframes in comparison to ground-truth.  For each sign $\mathcal{S}_{k},~k=1,..,N_S$,  let us denote by $l_{S_k},~k=1,..,N_S$ the number of keyframes extracted manually by the Greek SL experts. 
Let us also denote by $l_{X_k}, ~k=1,..,N_S$ the number of the extracted keyframes for each one of the $\mathcal{S}_{k}$ signs by using any computational method. Then, to assess the computational method's ability to capture the complexity of the signs in our dataset, we used the following $l_1$-based metric, i.e.:
\begin{equation}
C_S=\frac{1}{N_S}\sum_{k=1}^{N_S}|1-\frac{l_{X_k}}{l_{S_k}}|
\end{equation}

Note that, for each technique, the above defined metric quantifies the mean relative percentage in selecting the same number of keyframes with the annotator in each gloss. The obtained results in terms of the above defined metric versus $R_c$, are shown in Fig. \ref{fig:delta_recall_f2}(e). It is evident that the proposed figure of merit captures in a more effective way the sign's complexity than the other techniques do, achieving the lowest mean relative error.

\subsection{Human-Based Evaluation}
\label{subsec:HumanBased}
In this experiment we evaluated the intelligibility of the extracted summaries by using human SL experts.
We included the human generated summaries to evaluate subjectivity.
The evaluators were given only short videos depicting isolated signs and not the continuous sentences they were extracted from, because the latter are generally more easily interpretable due to their context. Context may lead to more concise summaries, but we left it for future work. 
More specifically, the extracted summaries, were used for the reconstruction of the original videos by repeating every keyframe, until the occurrence of the next one, so that the reconstructed video would have approximately the same duration with the original one. Those reconstructed videos, were evaluated by four expert interpreters in Greek SL who graded the depicted signs as, understandable (2), semi-understandable (1) and non-understandable (0).

Semi-understandable were those videos where the phonemic structure of the sign was not complete. In these videos a path movement or an internal (finger) movement was missing or an important handshape was missing. However, the lexical meaning was clear for the observer. The results of the human-based evaluation in approximately 500 signs of the dataset are shown in Table \ref{tab:human_based}. The superiority of the proposed technique is evident. 
The harmonic curvature outperforms the 2D curvature, but both the 2D and 3D curvature based techniques have almost the same performance. This suggests that using 3D curvature as a figure of merit may not be the best approach for solving the problem.
Finally, from Table \ref{tab:human_based} is clear that the 2-D $t$-parameterized curvature outperforms its $s$-parameterized counterpart thus validating our proposition.

\begin{table}[!ht]
\centering
\small
\caption{Human Based Evaluation: Proportion of glosses characterized from SL experts, as Understandable, Semi-Understandable and Non-Understandable from their keyframes}\label{tab:human_based}
		\newcolumntype{C}{>{\centering\arraybackslash}X}
  \begin{tabularx}{\linewidth}{CCCCCCCC}
\toprule
\multirow{2}{*}{} && \multicolumn{6}{c}{\textbf{Techniques}}\\
                  && Ground \nolinebreak Truth & $M(t)$ & $K_{2D}(t)$  & $\kappa_{3D}(s)$ \cite{geetha2013dynamic} & \multicolumn{1}{l}{$\kappa_{2D}(s)$  \cite{geetha2013dynamic}} & Moments \nolinebreak \cite{Kosmopoulos2005} \\ \midrule
\multicolumn{2}{c}{Understandable}      & 0.598    &0.534& 0.512  &  0.487    & 0.490      & 0.426  \\ 
\multicolumn{2}{c}{Semi-Understandable}  & 0.254   &0.243& 0.254  &  0.248    & 0.238      & 0.272  \\ 
\multicolumn{2}{c}{Non-Understandable}  & 0.148    &0.223& 0.234  &   0.265   & 0.272      & 0.302  \\ \bottomrule		\end{tabularx}
\end{table}

\subsection{Gloss Classification}
\label{subsec:GlossClassification}
In \cite{sartinas20212}, the authors evaluated the effectiveness of various keyframe extraction techniques, in the gloss classification problem, including the summaries created by human sign language experts.  In this paper we add as a baseline the case of the full video, i.e., using all the original frames for classification.  To this end, we grouped the selected keyframes by each technique into glosses and used the available annotation of glosses (including the start and stop timestamps and the meaning of each gloss) to identify their meaning in the available data.We used the obtained by MediaPipe keypoints with the setup described in \cite{sartinas20212}.

The results, measured in terms of top-$N$ accuracy for $N$=1, 2, 5, and 10, are shown in Table \ref{tab:classification}. We consider a classification to be top-$N$ accurate if the true meaning of the gloss appears in the $N$ most probable classes. The results demonstrate that the keyframes extracted using our proposed criterion are more effective at identifying the meanings of glosses, as they achieve higher accuracy scores. These results are promising given the sizeo of the vocabulary, i.e., 387 possible classes.

Note also that despite the huge difference in compression, the keyframes selected by the SL professionals outperform the full frames baseline, indicating the quality of the ground truth.This result also suggests that the original videos contain many frames that are noisy, e.g., due to blur or due to intermediate non-semantically significant hand poses; such frames can be discarded by using properly selected keyframes. 

\begin{table}[!ht]
\centering
\caption{Evaluation in classification task in the keyframe skeletal features obtained from proposed and techniques in \cite{geetha2013dynamic,Kosmopoulos2005,sartinas20212}.\label{tab:classification}}
\scriptsize
		\newcolumntype{C}{>{\centering\arraybackslash}X}
		\begin{tabularx}{\linewidth}{CCCCCCCCC}
  \toprule
    \multirow{2}{*}{} & \multicolumn{8}{c}{\textbf{Techniques}}\\
                  & Ground Truth & Full Frames & $M(t)$ & $K_{3D}(t)$ & $K_{2D}(t)$  \cite{sartinas20212} & $\kappa_{3D}(s)$ \cite{geetha2013dynamic} & \multicolumn{1}{l}{$\kappa_{2D}(s)$ \cite{geetha2013dynamic}} & \mbox{Moments \cite{Kosmopoulos2005}} \\ 
            \midrule
Top-1   & 0.56  & 0.52  & \textbf{0.44}  &  0.42  & 0.43  & 0.39  & 0.39  & 0.38\\ 
Top-2   & 0.70  & 0.64  & \textbf{0.56}  &  0.53  & 0.54  & 0.51  & 0.51  & 0.51\\ 
Top-5   & 0.82  & 0.78  & \textbf{0.69}  &  0.66  & 0.68  & 0.62  & 0.64  & 0.64\\ 
Top-10  & 0.88  & 0.85  & \textbf{0.76}  &  0.73  & 0.73  & 0.69  & 0.73  & 0.73\\ 
\bottomrule
		\end{tabularx}
 
\end{table}

\subsection{Ablation Study}

How well do the keyframes capture the meaning of sign language glosses? In this experiment we are going to answer this question via an ablation study. 
Specifically, we are going to solve the gloss classification problem under three different settings, by employing a state-of-the-art deep neural network, namely; a spatiotemporal graph convolutional neural network (ST-GCNN) \cite{yan2018spatial}.

We used the following settings:
\begin{itemize}
    \item In the first setting, we fed all frames within a specific gloss to the deep neural network.
    \item In the second setting, we excluded the keyframes extracted by the annotators, because these yield the best results in the gloss classification experiment, and their bilateral ($\pm 5$ frames) neighboring frames.
    \item Finally, under the third setting, which we consider as the baseline one, we excluded from the input the same number of randomly selected keyframes and their bilateral neighbors.

\end{itemize}

In all cases, we interpolated every gloss to 150 frames to ensure that all glosses had the same temporal length.

The Top-1 and Top-5 obtained results are shown in Table \ref{tab:st-gcn}. As it is clear the interpolation is not effective when keyframes are missing, leading to a significant drop in classification accuracy. This demonstrates the importance of keyframes in accurately capturing the meaning of sign language glosses. Our proposed technique for keyframe extraction, which utilizes both hand keypoints from mediapipe as features, seems to be effective in this regard.

\begin{table}[!htt] 
\caption{Evaluation of the ground truth keyframes' importance in classification's accuracy using ST-GCN}
\newcolumntype{C}{>{\centering\arraybackslash}X}
\centering
\resizebox{0.9\linewidth}{!}{%
\begin{tabularx}{\linewidth}{lCCCCC} 
\toprule
Accuracy &\textbf{Full Frames}	& \textbf{Exclude Random}	& \textbf{Exclude  GT  Keyframes} & \textbf{Exclude  $K_{2D}(t)$  Keyframes} & \textbf{Exclude  $M(t)$  Keyframes}\\
\midrule
Top-1   & 0.67  & 0.66  & 0.52  & 0.56  & 0.56\\
Top-2   & 0.80  & 0.79  & 0.64  & 0.72  & 0.69\\
Top-5   & 0.88  & 0.88  & 0.77  & 0.84  & 0.80\\
Top-10  & 0.93 & 0.92  & 0.82  & 0.88  & 0.88 \\
\bottomrule
\end{tabularx}}
\label{tab:st-gcn}

\end{table}
\unskip

\section{Conclusions}\label{sec:conclusion}

In this paper, we proposed a new method for identifying keyframes in videos by using $t$-parameterized curvature and torsion of the 3-D feature motion extracted from each frame. The proposed figure is the harmonic mean of $t$-parameterized curvature and torsion in the case of non-planar 3-D motion of the feature and $t$-parameterized curvature in the planar case, which is used to determine which frames are the most informative. We applied our keyframe extraction method to sign language videos, considering that the moving feature of importance is the signer's dominant wrist. We evaluated the proposed feature using objective measures with ground-truth keyframe annotations, human-based evaluations of understanding, and gloss classification in Greek Sign Language videos. The results of these experiments were promising.

\section{Acknowledgments}
The research was co-financed by Greece and EU through the Operational Program ``Human Resource Development, Education and Lifelong Learning”, 2014-2020, within the framework of the Action ”Strengthening human resources through the implementation of doctoral research - Action 2: IKY grant program for PhD candidates of Greek universities''.

\bibliography{references}

\begin{thebibliography}{10}
\providecommand{\url}[1]{#1}
\csname url@samestyle\endcsname
\providecommand{\newblock}{\relax}
\providecommand{\bibinfo}[2]{#2}
\providecommand{\BIBentrySTDinterwordspacing}{\spaceskip=0pt\relax}
\providecommand{\BIBentryALTinterwordstretchfactor}{4}
\providecommand{\BIBentryALTinterwordspacing}{\spaceskip=\fontdimen2\font plus
\BIBentryALTinterwordstretchfactor\fontdimen3\font minus
  \fontdimen4\font\relax}
\providecommand{\BIBforeignlanguage}[2]{{%
\expandafter\ifx\csname l@#1\endcsname\relax
\typeout{** WARNING: IEEEtran.bst: No hyphenation pattern has been}%
\typeout{** loaded for the language `#1'. Using the pattern for}%
\typeout{** the default language instead.}%
\else
\language=\csname l@#1\endcsname
\fi
#2}}
\providecommand{\BIBdecl}{\relax}
\BIBdecl

\bibitem{nikolaraizi2013}
M.~Nikolaraizi, I.~Vekiri, and S.~R. Easterbrooks, ``Investigating deaf
  students use of visual multimedia resources in reading comprehension,''
  \emph{American Annals of the Deaf}, vol. 157, no.~5, p. 458–473, 2013.

\bibitem{Rochan2018}
M.~Rochan, L.~Ye, and Y.~Wang, ``Video summarization using fully convolutional
  sequence networks,'' \emph{ArXiv}, vol. abs/1805.10538, 2018.

\bibitem{Guan2012}
G.~{Guan}, Z.~{Wang}, K.~{Yu}, S.~{Mei}, M.~{He}, and D.~{Feng}, ``Video
  summarization with global and local features,'' in \emph{2012 IEEE
  International Conference on Multimedia and Expo Workshops}, July 2012, pp.
  570--575.

\bibitem{Shao2009}
L.~{Shao} and L.~{Ji}, ``Motion histogram analysis based key frame extraction
  for human action/activity representation,'' in \emph{2009 Canadian Conference
  on Computer and Robot Vision}, May 2009, pp. 88--92.

\bibitem{Crasborn2001}
O.~Crasborn, ``Phonetic implementation of phonological categories in sign
  language of the netherlands,'' \emph{Sign Language \& Linguistics, Utrecht:
  Landelijke Onderzoeksschool Taalwetenschap}, vol.~5, no.~1, 2001.

\bibitem{Pfau2012}
R.~Pfau, M.~Steinbach, and B.~Woll, ``Sign language: An international
  handbook,'' \emph{Berlin: De Gruyter Mouton}, 2012.

\bibitem{Brentari1998}
B.~D., ``A prosodic model of sign language phonology,'' \emph{Cambridge, MA:
  MIT Press}, 1998.

\bibitem{sartinas20212}
E.~G. Sartinas, E.~Z. Psarakis, K.~Antzakas, and D.~I. Kosmopoulos, ``A 2-d
  wrist motion based sign language video summarization,'' \emph{In Proceedings
  of BMVC}, 2021.

\bibitem{Cernekova2002}
Z.~\={}Cernekov\'{a}, C.~Nikou, and I.~Pitas, ``Entropy metrics used for video
  summarization,'' in \emph{Proceedings of the 18th Spring Conference on
  Computer Graphics}, New York, NY, USA, 2002, pp. 73--82.

\bibitem{Chasanis2008}
V.~Chasanis, A.~Likas, and N.~Galatsanos, ``Efficient video shot summarization
  using an enhanced spectral clustering approach,'' in \emph{Artificial Neural
  Networks - ICANN}, V.~K{\r{u}}rkov{\'a}, R.~Neruda, and J.~Koutn{\'i}k,
  Eds.\hskip 1em plus 0.5em minus 0.4em\relax Springer, 2008, pp. 847--856.

\bibitem{Agyeman2019}
R.~{Agyeman}, R.~{Muhammad}, and G.~S. {Choi}, ``Soccer video summarization
  using deep learning,'' in \emph{2019 IEEE Conference on Multimedia
  Information Processing and Retrieval (MIPR)}, March 2019.

\bibitem{Lai2016}
{Po Kong Lai}, M.~{Décombas}, K.~{Moutet}, and R.~{Laganière}, ``Video
  summarization of surveillance cameras,'' in \emph{2016 13th IEEE
  International Conference on Advanced Video and Signal Based Surveillance
  (AVSS)}, Aug 2016, pp. 286--294.

\bibitem{Zhang2016}
K.~Zhang, W.-L. Chao, F.~Sha, and K.~Grauman, ``Video summarization with long
  short-term memory,'' in \emph{ECCV}, 2016.

\bibitem{Yang2015}
H.~Yang, B.~Wang, S.~Lin, D.~Wipf, M.~Guo, and B.~Guo, ``Unsupervised
  extraction of video highlights via robust recurrent auto-encoders,'' in
  \emph{Proceedings of the 2015 IEEE International Conference on Computer
  Vision (ICCV)}, ser. ICCV '15, 2015, pp. 4633--4641.

\bibitem{Mahasseni2017}
B.~{Mahasseni}, M.~{Lam}, and S.~{Todorovic}, ``Unsupervised video
  summarization with adversarial lstm networks,'' in \emph{2017 IEEE Conference
  on Computer Vision and Pattern Recognition (CVPR)}, July 2017, pp.
  2982--2991.

\bibitem{AGRAFIOTIS2006}
D.~Agrafiotis, N.~Canagarajah, D.~R. Bull, J.~Kyle, H.~Seers, and M.~Dye, ``A
  perceptually optimised video coding system for sign language communication at
  low bit rates,'' \emph{Signal Processing: Image Communication}, vol.~21,
  no.~7, pp. 531 -- 549, 2006.

\bibitem{Saxe2002}
D.~M. {Saxe} and R.~A. {Foulds}, ``Robust region of interest coding for
  improved sign language telecommunication,'' \emph{IEEE Transactions on
  Information Technology in Biomedicine}, vol.~6, no.~4, Dec 2002.

\bibitem{Tran2014}
J.~J. Tran, B.~Flowers, E.~A. Risken, R.~E. Ladner, and J.~O. Wobbrock,
  ``Analyzing the intelligibility of real-time mobile sign language video
  transmitted below recommended standards,'' in \emph{Proceedings of the 16th
  International ACM SIGACCESS Conference on Computers \& Accessibility}, ser.
  ASSETS '14.\hskip 1em plus 0.5em minus 0.4em\relax New York, NY, USA: ACM,
  2014, pp. 177--184.

\bibitem{Kosmopoulos2005}
D.~I. {Kosmopoulos}, A.~{Doulamis}, and N.~{Doulamis}, ``Gesture-based video
  summarization,'' in \emph{IEEE International Conference on Image Processing
  2005}, vol.~3, Sep. 2005, pp. III--1220.

\bibitem{geetha2013dynamic}
M.~Geetha and P.~Aswathi, ``Dynamic gesture recognition of indian sign language
  considering local motion of hand using spatial location of key maximum
  curvature points,'' in \emph{2013 IEEE Recent Advances in Intelligent
  Computational Systems (RAICS)}.\hskip 1em plus 0.5em minus 0.4em\relax IEEE,
  2013, pp. 86--91.

\bibitem{8846585}
D.~Guo, W.~Zhou, A.~Li, H.~Li, and M.~Wang, ``Hierarchical recurrent deep
  fusion using adaptive clip summarization for sign language translation,''
  \emph{IEEE Transactions on Image Processing}, vol.~29, pp. 1575--1590, 2020.

\bibitem{wang2018temporal}
L.~Wang, Y.~Xiong, Z.~Wang, Y.~Qiao, D.~Lin, X.~Tang, and L.~Van~Gool,
  ``Temporal segment networks for action recognition in videos,'' \emph{IEEE
  transactions on pattern analysis and machine intelligence}, vol.~41, no.~11,
  pp. 2740--2755, 2018.

\bibitem{zhu2016key}
W.~Zhu, J.~Hu, G.~Sun, X.~Cao, and Y.~Qiao, ``A key volume mining deep
  framework for action recognition,'' in \emph{Proceedings of the IEEE
  conference on computer vision and pattern recognition}, 2016, pp. 1991--1999.

\bibitem{chalmers2003visual}
A.~Chalmers, K.~Cater, and D.~Maflioli, ``Visual attention models for producing
  high fidelity graphics efficiently,'' in \emph{Proceedings of the 19th spring
  conference on Computer graphics}, 2003, pp. 39--45.

\bibitem{dilsizian2014}
M.~Dilsizian, P.~Yanovich, S.~Wang, C.~Neidle, and D.~Metaxas, ``A new
  framework for sign language recognition based on 3{D} handshape
  identification and linguistic modeling,'' in \emph{Proceedings of the Ninth
  International Conference on Language Resources and Evaluation ({LREC}'14)},
  May 2014, pp. 1924--1929.

\bibitem{ferger1931nature}
W.~F. Ferger, ``The nature and use of the harmonic mean,'' \emph{Journal of the
  American Statistical Association}, vol.~26, no. 173, 1931.

\bibitem{Lugaresi2019}
\BIBentryALTinterwordspacing
C.~L. et~all, ``Mediapipe: {A} framework for building perception pipelines,''
  \emph{CoRR}, 2019. [Online]. Available: \url{http://arxiv.org/abs/1906.08172}
\BIBentrySTDinterwordspacing

\bibitem{Fink}
E.~Fink and H.~S. Gandhi, ``Important extrema of time series,''
  \emph{https://www.cs.cmu.edu/eugene/research/full/important-extrema.pdf},
  2004.

\bibitem{yan2018spatial}
S.~Yan, Y.~Xiong, and D.~Lin, ``Spatial temporal graph convolutional networks
  for skeleton-based action recognition,'' in \emph{Proceedings of the AAAI
  conference on artificial intelligence}, vol.~32, no.~1, 2018.

\end{thebibliography}
\bibliographystyle{IEEEtran}

\end{document}